\documentclass[twocolumn,conference]{IEEEtran}
\usepackage[T1]{fontenc}
\usepackage[latin9]{inputenc}
\usepackage{color}
\usepackage{array}
\usepackage{rotating}
\usepackage{textcomp}
\usepackage{multirow}
\usepackage{graphicx}
\usepackage[unicode=true,
 bookmarks=true,bookmarksnumbered=true,bookmarksopen=true,bookmarksopenlevel=1,
 breaklinks=false,pdfborder={0 0 0},backref=false,colorlinks=false]
 {hyperref}
\hypersetup{pdftitle={Your Title},
 pdfauthor={Your Name},
 pdfborderstyle={},pdfpagelayout=OneColumn,pdfnewwindow=true,pdfstartview=XYZ,plainpages=false}
\usepackage{breakurl}

\makeatletter

\providecommand{\tabularnewline}{\\}

\usepackage{cite}

\makeatother

\begin{document}

\title{Identification of Patterns of Cognitive Impairment for Early Detection
of Dementia}

\author{Anusha A. S., Uma Ranjan,  Medha Sharma, and Siddharth Dutt\\
 \IEEEauthorblockA{\noindent {\footnotesize{}{}{}Centre for Brain Research, Indian
Institute of Science Bengaluru, India }\\
 {\footnotesize{}{}{}e-mail: anushas@iisc.ac.in, medhasharma@iisc.ac.in,
siddharthd@iisc.ac.in, uma@iisc.ac.in}}}
\maketitle
\begin{abstract}
Early detection of dementia is crucial to devise effective interventions.
Comprehensive cognitive tests, while being the most accurate means
of diagnosis, are long and tedious, thus limiting their applicability
to a large population, especially when periodic assessments are needed.
The problem is compounded by the fact that people have differing patterns
of cognitive impairment as they progress to different forms of dementia.
This paper presents a novel scheme by which individual-specific patterns
of impairment can be identified and used to devise personalized tests
for periodic follow-up. Patterns of cognitive impairment are initially
learned from a population cluster of combined normals and MCIs, using
a set of standardized cognitive tests. Impairment patterns in the
population are identified using a 2-step procedure involving an ensemble
wrapper feature selection followed by cluster identification and analysis.
These patterns have been shown to correspond to clinically accepted
variants of MCI, a prodrome of dementia. The learned clusters of patterns
can subsequently be used to identify the most likely route of cognitive
impairment, even for pre-symptomatic and apparently normal people.
Baseline data of 24,000 subjects from the NACC database was used for
the study.

\end{abstract}

\IEEEpeerreviewmaketitle{}

\section{Introduction}

Dementia is a neurodegenerative disorder with a high social burden.
Current diagnostic techniques do not recognize the onset of dementia
until it is too late for effective intervention. On the other hand,
the actual onset of the disease occurs many decades before symptoms
present themselves. Hence, it is important to detect it early, in
order to design effective interventions.

An early stage of dementia is Mild Cognitive Impairment (MCI) which
is preceded by a pre-symptomatic silent stage when the mechanism of
disease has started, but there are no noticeable symptoms. It is thus
desirable for even apparently normal individuals to undergo regular
check-ups to enable early detection of MCI.

Cognitive testing has been shown to be most effective in identifying
MCI \cite{darby2002mild}. Comprehensive cognitive testing batteries
are large and take a long time to administer. This leads to patient
fatigue, which in turn can cause poor performance on the test, leading
to false negatives. When longitudinal studies are undertaken, the
length of the tests also increases the cost of the study and limits
the percentage of the population that can be enrolled. Hence, there
is interest in shortening the length of the battery, especially when
used for periodic check-ups.

Recent clinical studies have shown that MCI is a clinically heterogeneous
condition with multiple subtypes \cite{Petersen2004}, with each corresponding
to different forms of dementia like Alzheimer's Disease, Lewy body
dementia, Vascular dementia or Frontotemporal dementia\cite{Subramanyam2016}.
These subtypes have different risk factors and prognosis.

Automated methods for early detection of dementia belong to two main
categories: (i) Distinguishing between normal and MCIs using classification
schemes\cite{Fraser2019,Kang2019} and (ii) Clustering a population
of known MCI to distinguish the various subtypes. These include Ward's
minimum variance clustering \cite{Edmonds2015,Clark2013,Delano-Wood2009},
k-means\cite{Libon2010}, Latent Profile Analysis \cite{Hanfelt2018,Eppig2017,McGuinness2015},
or Partially Ordered Subsets \cite{Tatsuoka2013}. 

However, these methods cannot be used for people in the pre-symptomatic
stage. In this work, we present a scheme by which we cluster a population
to identify patterns of cognitive impairment, irrespective of their
condition. This is done by a 2-step procedure where we first perform
an ensemble feature selection to distinguish between normals and MCIs,
without considering subtypes of MCI. A second step automatically clusters
the population using the selected features. Statistical analysis of
the clusters yields specific cognitive impairment patterns within
each cluster. These patterns, which are found to correspond to clinically
accepted subtypes of MCI, can be used to arrive at a shortened battery
for different sub-classes of the population. Section II describes
the details of the methodology used in the study. The study results
are discussed in Section III and conclusions are presented in Section
IV.

\section{Methodology}

\subsection{Dataset Description}

The study population comprised of 24,000 subjects, with equal numbers
of normals and MCIs (9937 Males and 14063 females; age (mean \textpm{}
SD): 71.4 \textpm{} 8.1 years), selected from the NACC Uniform Data
Set \cite{Beekly2007} gathered prospectively from approximately 32
Alzheimer's Disease Research Centers (ADRCs) between September 2005
and August 2019. The global rating of CDR\textsuperscript{\textregistered{}}\textit{Dementia
Staging Instrument} was used to determine the cognitive status (CN
or MCI) of subjects. 68 attributes obtained from 11 standardized neuropsychological
tests viz., Mini-Mental State Examination (MMSE), Logical memory (LM),
Copy of Benson Figure (BF), Forward and backward Digit Span (DS),
Category Fluency (CF), Trail Making Test (TMTA and TMTB), Wechsler
Adult Intelligence Scale (WAIS), Boston Naming Test (BNT), Verbal
Fluency (VF), Montreal Cognitive Assessment (MoCA), and Multilingual
Naming Test (MINT) were used as measures of cognitive functioning.
These features were identified as indicators of 6 cognitive domains
viz., orientation, memory, attention, visuospatial skills, language
and executive function. A detailed description of the neuropsychological
test battery can be found in \cite{Weintraub2009,Weintraub2018}.

\vspace{-0.1cm}

\subsection{Ensemble Feature Selection}

Feature selection is an important pre-processing step in data analysis
\cite{Saeys2007}, and is all the more important in the context of
the proposed scheme as the t-Distributed Stochastic Neighbor Embedding
(t-SNE) technique is used for visualization, where distances in large
dimensional spaces tend to be unreliable \cite{Colange2019}. An ensemble
wrapper approach with the union of features of multiple classifiers
was used, which has been shown to be be unbiased by any single classifier
\cite{Alvarez-Estevez2011}. Seven classification algorithms were
employed : Linear Discriminant Analysis (LDA), Quadratic Discriminant
Analysis (QDA), Na\"{i}ve Bayes (NB), \textit{k} Nearest Neighbor
(\textit{k}NN), Support Vector Machine with linear (SLI) and radial
basis function (SRBF) kernels, and Random Forest (RF). The 10-fold
cross-validation accuracy was used as the performance index. The final
feature subset was selected using the consensus strategy: \textit{Each
feature of the final feature subset is used by at least one of the
classifiers while performing wrapper based classification using k-fold
cross-validation}.

\vspace{-0.1cm}

\subsection{Identification of Population Subsets}

The identification of subsets within the population was performed
in 2 steps. In the first step, the sample space of feature vectors
which were obtained by ensemble wrapper selection was reduced to a
2-D manifold using the t-SNE projection. t-SNE is known to preserve
local structures of the data while converting from higher dimensions
to lower dimensions. However, it does not preserve information about
the large scale structure \cite{Maaten2008}. Therefore, it is ideally
suited to identify samples that are close to each other. Samples that
are far away from each other cannot be directly compared using the
t-SNE map, although they can be treated as separate subsets and analyzed
independently.

Population clusters were extracted from the t-SNE plot using a region-growing
segmentation based on morphological reconstruction \cite{Vincent1992MorphologicalGR}.
The marker image was a single pixel chosen in the interior of each
cluster. The mask image was a pixel array derived from the t-SNE co-ordinates.
The default 8-connectivity was used for the reconstruction.

\section{Results and Discussions}

\begin{table*}[t]
\protect\protect\caption{a summary of the ensemble wrapper-based feature selection scheme\label{tab:1}}

\begin{centering}
\begin{tabular}{|>{\centering}m{0.1cm}|>{\centering}m{0.5cm}|>{\centering}m{0.3cm}|>{\centering}m{0.54cm}|>{\centering}m{0.54cm}|>{\centering}m{0.54cm}|>{\centering}m{0.54cm}|>{\centering}m{0.54cm}|>{\centering}m{0.54cm}|>{\centering}m{0.54cm}|>{\centering}m{0.1cm}|>{\centering}m{0.5cm}|>{\centering}m{0.3cm}|>{\centering}m{0.54cm}|>{\centering}m{0.54cm}|>{\centering}m{0.54cm}|>{\centering}m{0.54cm}|>{\centering}m{0.54cm}|>{\centering}m{0.54cm}|>{\centering}m{0.54cm}|}
\hline 
\multirow{4}{0.1cm}{\begin{turn}{90}
{\footnotesize{}Test}
\end{turn}} & \multirow{4}{0.5cm}{\begin{turn}{90}
{\footnotesize{}ID} 
\end{turn}} & \multirow{4}{0.3cm}{\begin{turn}{90}
{\footnotesize{}CG} 
\end{turn}} & \multicolumn{7}{c|}{{\footnotesize{}{}{}Classifier}} & \multirow{4}{0.1cm}{\begin{turn}{90}
{\footnotesize{}Test} 
\end{turn}} & \multirow{4}{0.5cm}{\begin{turn}{90}
{\footnotesize{}ID} 
\end{turn}} & \multirow{4}{0.3cm}{\begin{turn}{90}
{\footnotesize{}CG} 
\end{turn}} & \multicolumn{7}{c|}{{\footnotesize{}{}{}Classifier}}\tabularnewline
\cline{4-10} \cline{14-20} 
 &  &  & {\footnotesize{}{}{}LDA}  & {\footnotesize{}{}{}QDA}  & \textit{\footnotesize{}{}{}k}{\footnotesize{}{}{}NN}  & {\footnotesize{}{}{}NB}  & {\footnotesize{}{}{}SLI}  & {\footnotesize{}{}{}SRBF}  & {\footnotesize{}{}{}RF}  &  &  &  & {\footnotesize{}{}{}LDA}  & {\footnotesize{}{}{}QDA}  & \textit{\footnotesize{}{}{}k}{\footnotesize{}{}{}NN}  & {\footnotesize{}{}{}NB}  & {\footnotesize{}{}{}SLI}  & {\footnotesize{}{}{}SRBF}  & {\footnotesize{}{}{}RF}\tabularnewline
\cline{4-10} \cline{14-20} 
 &  &  & \multicolumn{7}{c||}{{\footnotesize{}{}{}Cross validation accuracy (\%)}} &  &  &  & \multicolumn{7}{c|}{{\footnotesize{}{}{}Cross validation accuracy (\%)}}\tabularnewline
\cline{4-10} \cline{14-20} 
 &  &  & {\footnotesize{}{}{}73.98}  & {\footnotesize{}{}{}71.26}  & {\footnotesize{}{}{}73.82}  & {\footnotesize{}{}{}71.76}  & {\footnotesize{}{}{}73.42}  & {\footnotesize{}{}{}76.21}  & \multicolumn{1}{>{\centering}m{0.54cm}||}{{\footnotesize{}{}{}73.94}} &  &  &  & {\footnotesize{}{}{}73.98}  & {\footnotesize{}{}{}71.26}  & {\footnotesize{}{}{}73.82}  & {\footnotesize{}{}{}71.76}  & {\footnotesize{}{}{}73.42}  & {\footnotesize{}{}{}76.21}  & {\footnotesize{}{}{}73.94}\tabularnewline
\hline 
\multirow{3}{0.1cm}{\begin{turn}{90}
{\footnotesize{}MMSE}
\end{turn}} & \textbf{\textcolor{red}{\footnotesize{}{}{}F1}}{\footnotesize{}{}
}  & {\footnotesize{}{}{}O}  & {*}  &  & {*}  & {*}  & {*}  & {*}  & {*}  & \multirow{7}{0.1cm}{\begin{turn}{90}
{\footnotesize{}VF}
\end{turn}} & {\footnotesize{}{}{}F35}  & {\footnotesize{}{}{}L}  &  &  &  &  &  &  & \tabularnewline
\cline{2-10} \cline{12-20} 
 & \textbf{\textcolor{red}{\footnotesize{}{}{}F2}}{\footnotesize{}{}
}  & {\footnotesize{}{}{}O}  &  & {*}  &  &  & {*}  & {*}  &  &  & {\footnotesize{}{}{}F36}  & {\footnotesize{}{}{}L}  &  &  &  &  &  &  & \tabularnewline
\cline{2-10} \cline{12-20} 
 & \textbf{\textcolor{red}{\footnotesize{}{}{}F3}}{\footnotesize{}{}
}  & {\footnotesize{}{}{}V}  & {*}  & {*}  &  &  & {*}  &  &  &  & \textbf{\textcolor{red}{\footnotesize{}{}{}F37}}{\footnotesize{}{}
}  & {\footnotesize{}{}{}L}  &  &  &  &  & {*}  &  & \tabularnewline
\cline{1-10} \cline{12-20} 
\multirow{8}{0.1cm}{\begin{turn}{90}
{\footnotesize{}LM}
\end{turn}} & \textbf{\textcolor{red}{\footnotesize{}{}{}F4}}{\footnotesize{}{}
}  & {\footnotesize{}{}{}M}  & {*}  &  & {*}  & {*}  & {*}  & {*}  & {*}  &  & \textbf{\textcolor{red}{\footnotesize{}{}{}F38}}{\footnotesize{}{}
}  & {\footnotesize{}{}{}L}  &  &  &  &  & {*}  &  & \tabularnewline
\cline{2-10} \cline{12-20} 
 & {\footnotesize{}{}{}F5}  & {\footnotesize{}{}{}M}  &  &  &  &  &  &  &  &  & {\footnotesize{}{}{}F39}  & {\footnotesize{}{}{}L}  &  &  &  &  &  &  & \tabularnewline
\cline{2-10} \cline{12-20} 
 & {\footnotesize{}{}{}F6}  & {\footnotesize{}{}{}M}  &  &  &  &  &  &  &  &  & \textbf{\textcolor{red}{\footnotesize{}{}{}F40}}{\footnotesize{}{}
}  & {\footnotesize{}{}{}L}  &  &  &  &  & {*}  &  & \tabularnewline
\cline{2-10} \cline{12-20} 
 & {\footnotesize{}{}{}F7}  & {\footnotesize{}{}{}M}  &  &  &  &  &  &  &  &  & {\footnotesize{}{}{}F41}  & {\footnotesize{}{}{}L}  &  &  &  &  &  &  & \tabularnewline
\cline{2-20} 
 & \textbf{\textcolor{red}{\footnotesize{}{}{}F8}}{\footnotesize{}{}
}  & {\footnotesize{}{}{}M}  &  &  &  &  & {*}  & {*}  &  & \multirow{21}{0.1cm}{\begin{turn}{90}
{\footnotesize{}MoCA} 
\end{turn}} & {\footnotesize{}{}{}F42}  & {\footnotesize{}{}{}V}  &  &  &  &  &  &  & \tabularnewline
\cline{2-10} \cline{12-20} 
 & {\footnotesize{}{}{}F9}  & {\footnotesize{}{}{}M}  &  &  &  &  &  &  &  &  & {\footnotesize{}{}{}F43}  & {\footnotesize{}{}{}V}  &  &  &  &  &  &  & \tabularnewline
\cline{2-10} \cline{12-20} 
 & {\footnotesize{}{}{}F10}  & {\footnotesize{}{}{}M}  &  &  &  &  &  &  &  &  & \textbf{\textcolor{red}{\footnotesize{}{}{}F44}}{\footnotesize{}{}
}  & {\footnotesize{}{}{}V}  & {*}  &  & {*}  &  &  &  & \tabularnewline
\cline{2-10} \cline{12-20} 
 & \textbf{\textcolor{red}{\footnotesize{}{}{}F11}}{\footnotesize{}{}
}  & {\footnotesize{}{}{}M}  &  &  & {*}  &  &  &  & {*}  &  & {\footnotesize{}{}{}F45}  & {\footnotesize{}{}{}V}  &  &  &  &  &  &  & \tabularnewline
\cline{1-10} \cline{12-20} 
\multirow{3}{0.1cm}{\begin{turn}{90}
{\footnotesize{}BF} 
\end{turn}} & \textbf{\textcolor{red}{\footnotesize{}{}{}F12}}{\footnotesize{}{}
}  & {\footnotesize{}{}{}V}  &  & {*}  &  &  & {*}  &  &  &  & \textbf{\textcolor{red}{\footnotesize{}{}{}F46}}{\footnotesize{}{}
}  & {\footnotesize{}{}{}V}  & {*}  &  &  &  &  &  & {*}\tabularnewline
\cline{2-10} \cline{12-20} 
 & \textbf{\textcolor{red}{\footnotesize{}{}{}F13}}{\footnotesize{}{}
}  & {\footnotesize{}{}{}M}  &  &  &  &  & {*}  &  &  &  & {\footnotesize{}{}{}F47}  & {\footnotesize{}{}{}L}  &  &  &  &  &  &  & \tabularnewline
\cline{2-10} \cline{12-20} 
 & {\footnotesize{}{}{}F14}  & {\footnotesize{}{}{}M}  &  &  &  &  &  &  &  &  & \textbf{\textcolor{red}{\footnotesize{}{}{}F48}}{\footnotesize{}{}
}  & {\footnotesize{}{}{}M}  &  &  &  &  & {*}  &  & \tabularnewline
\cline{1-10} \cline{12-20} 
\multirow{8}{0.1cm}{\begin{turn}{90}
{\footnotesize{}DS} 
\end{turn}} & \textbf{\textcolor{red}{\footnotesize{}{}{}F15}}{\footnotesize{}{}
}  & {\footnotesize{}{}{}A}  &  &  &  &  &  & {*}  &  &  & \textbf{\textcolor{red}{\footnotesize{}{}{}F49}}{\footnotesize{}{}
}  & {\footnotesize{}{}{}A}  & {*}  &  &  &  &  &  & \tabularnewline
\cline{2-10} \cline{12-20} 
 & {\footnotesize{}{}{}F16}  & {\footnotesize{}{}{}A}  &  &  &  &  &  &  &  &  & \textbf{\textcolor{red}{\footnotesize{}{}{}F50}}{\footnotesize{}{}
}  & {\footnotesize{}{}{}A}  &  &  & {*}  &  &  &  & {*}\tabularnewline
\cline{2-10} \cline{12-20} 
 & {\footnotesize{}{}{}F17}  & {\footnotesize{}{}{}E}  &  &  &  &  &  &  &  &  & {\footnotesize{}{}{}F51}  & {\footnotesize{}{}{}A}  &  &  &  &  &  &  & \tabularnewline
\cline{2-10} \cline{12-20} 
 & {\footnotesize{}{}{}F18}  & {\footnotesize{}{}{}E}  &  &  &  &  &  &  &  &  & {\footnotesize{}{}{}F52}  & {\footnotesize{}{}{}L}  &  &  &  &  &  &  & \tabularnewline
\cline{2-10} \cline{12-20} 
 & \textbf{\textcolor{red}{\footnotesize{}{}{}F19}}{\footnotesize{}{}
}  & {\footnotesize{}{}{}A}  &  &  &  &  &  & {*}  &  &  & {\footnotesize{}{}{}F53}  & {\footnotesize{}{}{}L}  &  &  &  &  &  &  & \tabularnewline
\cline{2-10} \cline{12-20} 
 & \textbf{\textcolor{red}{\footnotesize{}{}{}F20}}{\footnotesize{}{}
}  & {\footnotesize{}{}{}A}  &  &  &  &  & {*}  &  &  &  & \textbf{\textcolor{red}{\footnotesize{}{}{}F54}}{\footnotesize{}{}
}  & {\footnotesize{}{}{}M}  & {*}  &  & {*}  & {*}  & {*}  & {*}  & {*}\tabularnewline
\cline{2-10} \cline{12-20} 
 & {\footnotesize{}{}{}F21}  & {\footnotesize{}{}{}E}  &  &  &  &  &  &  &  &  & {\footnotesize{}{}{}F55}  & {\footnotesize{}{}{}M}  &  &  &  &  &  &  & \tabularnewline
\cline{2-10} \cline{12-20} 
 & \textbf{\textcolor{red}{\footnotesize{}{}{}F22}}{\footnotesize{}{}
}  & {\footnotesize{}{}{}E}  &  &  &  &  & {*}  &  &  &  & {\footnotesize{}{}{}F56}  & {\footnotesize{}{}{}M}  &  &  &  &  &  &  & \tabularnewline
\cline{1-10} \cline{12-20} 
\multirow{2}{0.1cm}{\begin{turn}{90}
{\footnotesize{}CF}
\end{turn}} & \textbf{\textcolor{red}{\footnotesize{}{}{}F23}}{\footnotesize{}{}
}  & {\footnotesize{}{}{}L}  &  & {*}  &  & {*}  &  & {*}  &  &  & \textbf{\textcolor{red}{\footnotesize{}{}{}F57}}{\footnotesize{}{}
}  & {\footnotesize{}{}{}O}  &  &  &  &  &  &  & {*}\tabularnewline
\cline{2-10} \cline{12-20} 
 & \textbf{\textcolor{red}{\footnotesize{}{}{}F24}}{\footnotesize{}{}
}  & {\footnotesize{}{}{}L}  &  &  & {*}  & {*}  & {*}  & {*}  & {*}  &  & {\footnotesize{}{}{}F58}  & {\footnotesize{}{}{}O}  &  &  &  &  &  &  & \tabularnewline
\cline{1-10} \cline{12-20} 
\multirow{6}{0.1cm}{\begin{turn}{90}
{\footnotesize{}TMT} 
\end{turn}} & \textbf{\textcolor{red}{\footnotesize{}{}{}F25}}{\footnotesize{}{}
}  & {\footnotesize{}{}{}A}  & {*}  &  &  &  & {*}  &  &  &  & \textbf{\textcolor{red}{\footnotesize{}{}{}F59}}{\footnotesize{}{}
}  & {\footnotesize{}{}{}O}  &  &  &  &  &  &  & {*}\tabularnewline
\cline{2-10} \cline{12-20} 
 & {\footnotesize{}{}{}F26}  & {\footnotesize{}{}{}A}  &  &  &  &  &  &  &  &  & \textbf{\textcolor{red}{\footnotesize{}{}{}F60}}{\footnotesize{}{}
}  & {\footnotesize{}{}{}O}  &  &  & {*}  &  &  &  & {*}\tabularnewline
\cline{2-10} \cline{12-20} 
 & {\footnotesize{}{}{}F27}  & {\footnotesize{}{}{}A}  &  &  &  &  &  &  &  &  & {\footnotesize{}{}{}F61}  & {\footnotesize{}{}{}O}  &  &  &  &  &  &  & \tabularnewline
\cline{2-10} \cline{12-20} 
 & \textbf{\textcolor{red}{\footnotesize{}{}{}F28}}{\footnotesize{}{}
}  & {\footnotesize{}{}{}E}  &  &  &  &  &  & {*}  &  &  & \textbf{\textcolor{red}{\footnotesize{}{}{}F62}}{\footnotesize{}{}
}  & {\footnotesize{}{}{}O}  &  &  & {*}  &  &  &  & \tabularnewline
\cline{2-20} 
 & {\footnotesize{}{}{}F29}  & {\footnotesize{}{}{}E}  &  &  &  &  &  &  &  & \multirow{6}{0.1cm}{\begin{turn}{90}
M{\footnotesize{}INT} 
\end{turn}} & {\footnotesize{}{}{}F63}  & {\footnotesize{}{}{}L}  &  &  &  &  &  &  & \tabularnewline
\cline{2-10} \cline{12-20} 
 & \textbf{\textcolor{red}{\footnotesize{}{}{}F30}}{\footnotesize{}{}
}  & {\footnotesize{}{}{}E}  &  &  &  &  & {*}  &  &  &  & {\footnotesize{}{}{}F64}  & {\footnotesize{}{}{}L}  &  &  &  &  &  &  & \tabularnewline
\cline{1-10} \cline{12-20} 
\begin{turn}{90}
{\footnotesize{}W} 
\end{turn} & \textbf{\textcolor{red}{\footnotesize{}{}{}F31}}{\footnotesize{}{}
}  & {\footnotesize{}{}{}E}  &  & {*}  &  &  &  & {*}  &  &  & {\footnotesize{}{}{}F65}  & {\footnotesize{}{}{}L}  &  &  &  &  &  &  & \tabularnewline
\cline{1-10} \cline{12-20} 
\begin{turn}{90}
 {\footnotesize{}B} 
\end{turn} & {\footnotesize{}{}{}F32}  & {\footnotesize{}{}{}L}  &  &  &  &  &  &  &  &  & {\footnotesize{}{}{}F66}  & {\footnotesize{}{}{}L}  &  &  &  &  &  &  & \tabularnewline
\cline{1-10} \cline{12-20} 
\multirow{2}{0.1cm}{\begin{turn}{90}
{\footnotesize{}VF} 
\end{turn}} & \textbf{\textcolor{red}{\footnotesize{}{}{}F33}}{\footnotesize{}{}
}  & {\footnotesize{}{}{}L}  &  &  &  & {*}  &  &  &  &  & \textbf{\textcolor{red}{\footnotesize{}{}{}F67}}{\footnotesize{}{}
}  & {\footnotesize{}{}{}L}  & {*}  &  &  &  &  &  & \tabularnewline
\cline{2-10} \cline{12-20} 
 & {\footnotesize{}{}{}F34}  & {\footnotesize{}{}{}L}  &  &  &  &  &  &  &  &  & {\footnotesize{}{}{}F68}  & {\footnotesize{}{}{}L}  &  &  &  &  &  &  & \tabularnewline
\hline 
\end{tabular}
\par\end{centering}

\vspace{0.1cm}

\textit{\footnotesize{}{}{}ID refers to Feature Identifier; CG indicates
the cognitive domain to which a feature belongs. A: Attention, E:
Executive function, L: Language, M: Memory, O: Orientation, V: Visuospatial;W
represents the WAIS test and B represents BNT; Features selected by
each classifier is indicated by `{*}'. The final subset of 33 features
is highligted in red.}{\footnotesize \par}

\vspace{-0.5cm}
\end{table*}

The results of the ensemble wrapper-based feature selection is presented
in TABLE \ref{tab:1}. The features chosen by the consensus strategy
are highlighted in red. It is seen that all the cognitive domains
are covered. This is expected, since at this stage, no distinction
has been made among the MCI subtypes.

\begin{figure}
\includegraphics[scale=0.38]{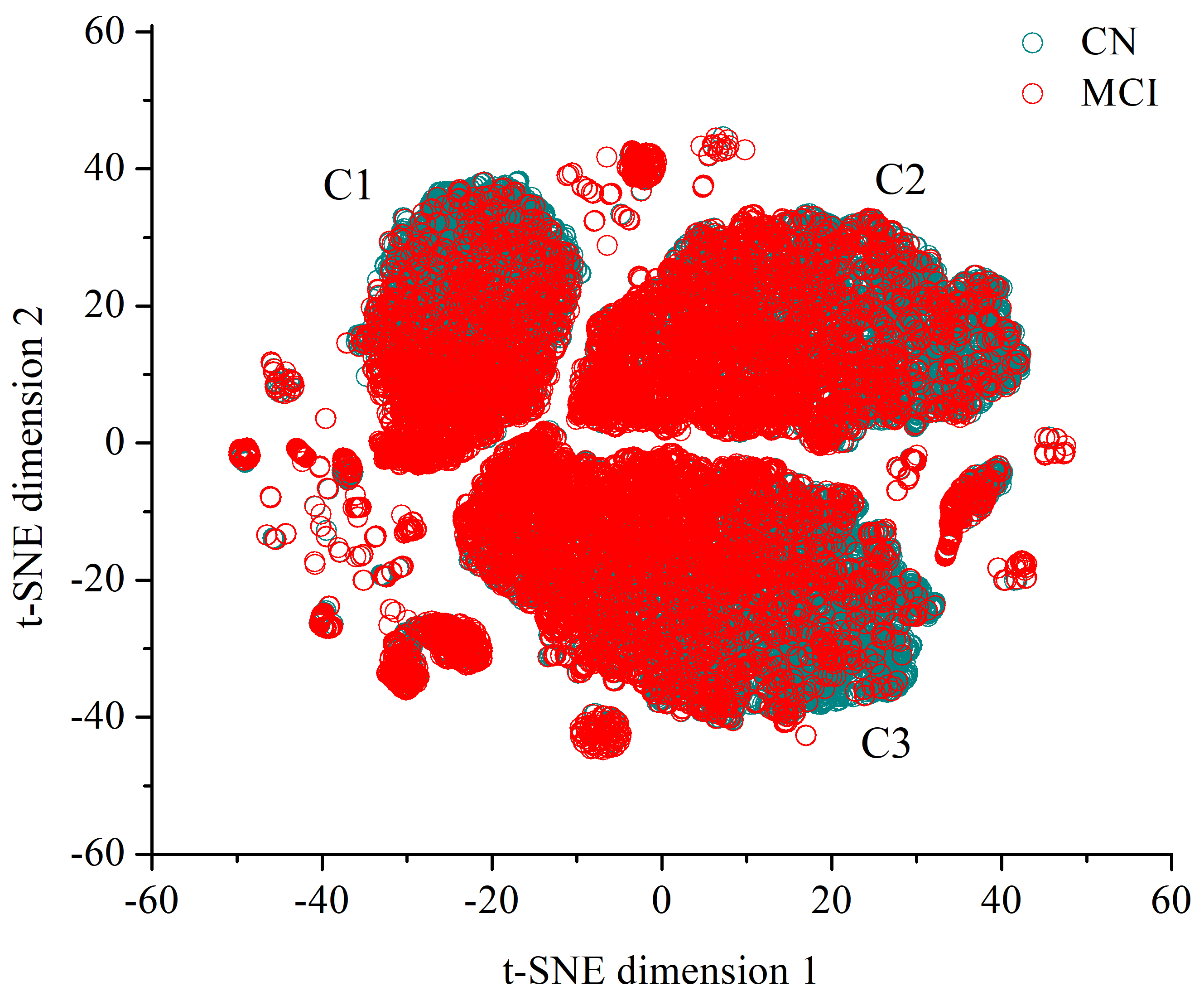}

\protect\protect\caption{2-D visualization of the dataset using t-SNE\label{fig:1}}

\vspace{-0.2cm}
\end{figure}

The results of the next stage, where the selected features are visualized
using a t-SNE plot is shown in Fig. \ref{fig:1}. This plot has been
gnerated from 24,000 instances using 33 features.

\begin{figure*}[tbh]
\includegraphics[scale=0.39]{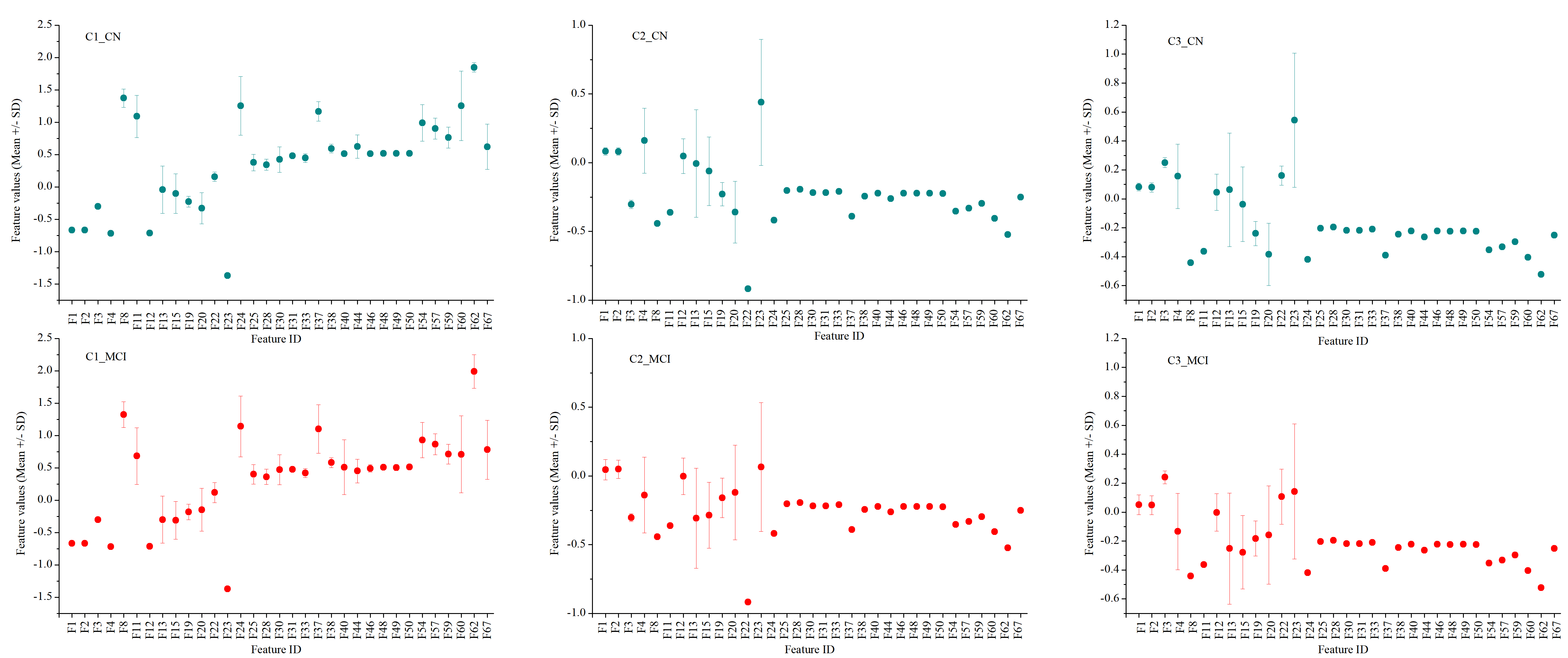}

\protect\protect\caption{Feature variations of CN and MCI populations within each subgroup\label{fig:2}}

\vspace{-0.4cm}
\end{figure*}

It is seen that the t-SNE plot identified 3 distinct clusters, in
line with the results in \cite{Libon2010}. The distribution of samples
across the subgroups, obtained after segmentation, is shown in TABLE
\ref{tab:2}. The statistical variation of the features within each
of the clusters is presented in Fig \ref{fig:2}. It is seen that
the distribution of features is very similar for the normals and MCIs
within a cluster, with most features showing no variations between
normals and MCI and a few showing minor variations. Moreover, the
MCIs in a cluster are more similar to the normals within a cluster
rather than to the MCIs of other clusters. This shows that MCI is
clearly a heterogeneous condition.

\begin{table}[tbh]
\protect\protect\caption{population distribution across subgroups \label{tab:2}}

\centering{}%
\begin{tabular}{|c|c|c|c|}
\hline 
Subset  & \# CN  & \# MCI  & Total\tabularnewline
\hline 
C1  & 752  & 1143  & 1895\tabularnewline
\hline 
C2  & 3660  & 3452  & 7112\tabularnewline
\hline 
C3  & 4437  & 4192  & 8629\tabularnewline
\hline 
\end{tabular}
\end{table}

The utility of the feature subset in differentiating CN and MCI samples
within each subgroup was further investigated separately, using statistical
techniques. Mann Whitney U test \cite{Sheskin2003} with a significance
level of $\alpha=0.05$ was performed on the CN and MCI data samples
within each subgroup. A non-parametric test like Mann Whitney is most
appropriate here, as most features do not follow a normal distribution.
For the subgroup C1, the test indicated that 27 out of 33 features
(all features except F1, F2, F3, F4, F15, and F31) exhibited statistically
significant differences between CN and MCI samples. Features F1, F2,
F4, F15, F23, F24, F25, F28, and F31 were found to show statistically
significant difference between the CN and MCI samples of C2, while
for C3, features F1, F2, F3, F4, F15, F23, F24, F25, F28, F30, and
F31 were chosen. However, statistical significance may not always
indicate an effect with a high impact. Therefore, the effect size
of the statistically significant differences was determined using
Cohen's d effect size \cite{cohen1992power}. The effect size was
computed for the features that showed significant differences between
the CN and MCI samples of each subgroup. Further, these effect sizes
were interpreted based on thresholds defined in \cite{cohen1992power},
i.e., |d|\ensuremath{\le}0.2 is `negligible' effect size, 0.2<|d|\ensuremath{\le}0.5
is `small', 0.5<|d|\ensuremath{\le}0.8 is `medium' and otherwise
`large'. A summary of features that indicated a large effect size
for each of the subgroups is shown in TABLE \ref{tab:3}

\begin{table}[tbh]
\vspace{-0.4cm}

\protect\protect\caption{features with large effect size \label{tab:3}}

\begin{centering}
\begin{tabular}{|>{\centering}p{1cm}|>{\raggedright}m{6.85cm}|}
\hline 
Subgroup  & \hspace{2cm}Features\tabularnewline
\hline 
C1  & F8 (M), F12 (V), F19 (A), F20 (A), F22 (E), F23 (L), F24 (L), F30
(E), F33 (L), F37 (L), F38 (L), F40 (L), F44 (V), F46 (V), F48 (M),
F49 (A), F50 (A), F54 (M), F57 (O), F59 (O), F60 (O), F62 (O), F67
(L)\tabularnewline
\hline 
C2  & F25 (A), F28 (E)\tabularnewline
\hline 
C3  & F3 (V), F25 (A), F28 (E)\tabularnewline
\hline 
\end{tabular}
\par\end{centering}

\smallskip{}

\begin{centering}
\textit{\footnotesize{}{}{}Cognitive domains are shown within brackets}{\footnotesize{} }
\par\end{centering}{\footnotesize \par}

\vspace{-0.2cm}
\end{table}

It is now seen that the different clusters show variations in a subset
of the domains, unlike the features obtained from the first stage
where all cognitive domains were represented. Features corresponding
to C1 shows variations in all cognitive domains. While C2 shows predominant
variations in attention and executive function domains, C3 indicates
variations in visuospatial, attention and executive function. Accordingly,
the MCI group in C1 can be characterized as amnestic multi-domain
MCI, while C2 and C3 can be categorized as non-amnestic multi-domain
MCI\cite{Subramanyam2016}.

Since the cluster contain both normals as well as MCIs, these sub-types
also correspond to the individuals diagnosed as normals. Thus,a person
currently classified as normal can also be associated with the closest
MCI sub-type and the possible route to cognitive impairment, should
he/she develop it.

By comparing the distribution of normals and MCIs within Fig \ref{fig:1},
it is seen that a gradation in the distribution of normals and MCIs
is seen within each cluster, with one end of each cluster corresponding
to a largely normal population, to the other end which consists of
largely MCI population. 

\vspace{-0.15cm}

\section{Conclusions}

We have proposed a novel scheme to identify the pattern of impairment
in different subsets of the population. These patterns of impairment
have been shown to correspond to clinically accepted routes to different
forms of dementia. This enables devising of shorter and cost-effective
cognitive battery that can be used for periodic check up of people
in normal and pre-symptomatic stage, and for regular follow-up in
longitudinal studies.

\vspace{-0.15cm}

\section*{Acknowlegment}

\begin{tiny}The NACC database is funded by NIA/NIH Grant U01 AG016976.
NACC data are contributed by the NIA-funded ADCs: P30 AG019610 (PI
Eric Reiman, MD), P30 AG013846 (PI Neil Kowall, MD), P30 AG062428-01
(PI James Leverenz, MD) P50 AG008702 (PI Scott Small, MD), P50 AG025688
(PI Allan Levey, MD, PhD), P50 AG047266 (PI Todd Golde, MD, PhD),
P30 AG010133 (PI Andrew Saykin, PsyD), P50 AG005146 (PI Marilyn Albert,
PhD), P30 AG062421-01 (PI Bradley Hyman, MD, PhD), P30 AG062422-01
(PI Ronald Petersen, MD, PhD), P50 AG005138 (PI Mary Sano, PhD), P30
AG008051 (PI Thomas Wisniewski, MD), P30 AG013854 (PI Robert Vassar,
PhD), P30 AG008017 (PI Jeffrey Kaye, MD), P30 AG010161 (PI David Bennett,
MD), P50 AG047366 (PI Victor Henderson, MD, MS), P30 AG010129 (PI
Charles DeCarli, MD), P50 AG016573 (PI Frank LaFerla, PhD), P30 AG062429-01(PI
James Brewer, MD, PhD), P50 AG023501 (PI Bruce Miller, MD), P30 AG035982
(PI Russell Swerdlow, MD), P30 AG028383 (PI Linda Van Eldik, PhD),
P30 AG053760 (PI Henry Paulson, MD, PhD), P30 AG010124 (PI John Trojanowski,
MD, PhD), P50 AG005133 (PI Oscar Lopez, MD), P50 AG005142 (PI Helena
Chui, MD), P30 AG012300 (PI Roger Rosenberg, MD), P30 AG049638 (PI
Suzanne Craft, PhD), P50 AG005136 (PI Thomas Grabowski, MD), P30 AG062715-01
(PI Sanjay Asthana, MD, FRCP), P50 AG005681 (PI John Morris, MD),
P50 AG047270 (PI Stephen Strittmatter, MD, PhD). \end{tiny} \vspace{-0.1cm}

\bibliographystyle{IEEEtran}
\bibliography{EMBC_2020_Edited}

\end{document}